# ANALOG SIGNAL PROCESSING APPROACH FOR COARSE AND FINE DEPTH ESTIMATION

Nihar Athreyas[1], Zhiguo Lai [2], Jai Gupta[2] and Dev Gupta[2]

[1]Department of Electrical and Computer Engineering, University of Massachusetts, Amherst, MA, USA
nathreya@umass.edu
[2]Newlans Inc., 43 Nagog Park, Acton, MA, USA
zlai@newlans.com

## ABSTRACT

*Imaging and Image sensors is a field that is continuously evolving. There are new products coming into the market every day. Some of these have very severe Size, Weight and Power constraints whereas other devices have to handle very high computational loads. Some require both these conditions to be met simultaneously. Current imaging architectures and digital image processing solutions will not be able to meet these ever increasing demands. There is a need to develop novel imaging architectures and image processing solutions to address these requirements. In this work we propose analog signal processing as a solution to this problem. The analog processor is not suggested as a replacement to a digital processor but it will be used as an augmentation device which works in parallel with the digital processor, making the system faster and more efficient. In order to show the merits of analog processing two stereo correspondence algorithms are implemented. We propose novel modifications to the algorithms and new imaging architectures which, significantly reduces the computation time.*

## 1. INTRODUCTION

The imaging and image sensor industry is going through a huge wave of change. Very soon there is going to be a great demand in the market for wearables like smart watches, headbands, glasses etc. Cameras will be an integral part of these devices. However these devices have severe Size, Weight and Power (SWaP) constraints. On the other hand companies are also trying to develop multi-megapixel sensors and there have been talks of developing gigapixel sensors for use in defence, space and medical applications. Collecting such huge amounts of data and processing it is not an easy task. There are a of lot emerging applications in the field of computer vision, biometric analysis, bio-medical imaging etc. which require ultra-high speed computations. Another area that is gaining traction is the use of stereo cameras, camera arrays and light-field cameras to perform computational imaging tasks. Current imaging architectures and digital image processing solutions will not work in all of these situations because they will not be able to handle the high computational loads and meet the SWaP requirements simultaneously. Hence there is a need for novel ideas and solutions that can address these requirements.

In this paper we reintroduce the concept of analog image processing and present it as a solution to the above problems of reducing SWaP and high computational load. Generally the use of the term analog image processing has been restricted to film photography or optical processing. We are using neither of these approaches but we are performing analog signal processing by considering the image data to be a continuous stream of analog voltage values.

In order to show the advantages of analog processing we chose the problem of image alignment in stereo cameras. Image pairs captured from the stereo cameras can be used for a variety of purposes like constructing disparity and depth maps, refocusing, to simulate the effect of optical zoom etc. Image alignment can be used for stitching images to create panoramas, for video stabilization, scene summarization etc. Whatever the application, one of the most important steps in stereo image processing is to find correspondence between the points in the two images which represents the same 3D point in the scene. This has been an active area of research for many years now and there are a lot of stereo correspondence algorithms that have been developed. However some of these algorithms are either slow or have poor performance in the presence of noise or low light. The reason for these algorithms being slow is the very high computational requirement. In this work we discuss the implementation of two stereo correspondence algorithms.

The first algorithm is a patch based stereo correspondence algorithm. Normalized Cross Correlation (NCC) is chosen as the similarity measure for this approach. NCC is very robust to noise and changes in the image intensity values but it is not preferred because of its high computational intensity. The second algorithm is a per-pixel algorithm which produces a finer disparity map than the patch based approach. Sum Absolute Differences (SAD) is used as a similarity measure for this algorithm.

In this paper we propose novel modifications to the algorithms which improves the computational speed without compromising the performance and also making it efficiently implementable in hardware. We also propose new circuit architectures that can be used to implement the modified NCC algorithm and modified SAD algorithm in the analog domain. The analog domain implementation provides further speedup in computation and has lower power consumption than a digital implementation.

The organization of the paper is as follows. Section 2 discusses the patch based stereo correspondence algorithm, proposed modifications, hardware architecture and some experimental results. Section 3 discusses the per-pixel algorithm, proposed modifications, hardware architecture and experimental results. The work is concluded in section 4.

# 2. PATCH BASED STEREO CORRESPONDENCE ALGORITHM

All stereo correspondence algorithms can be broadly classified into intensity based algorithms and feature based algorithms. In feature based algorithms features such as edges and contours are extracted from both stereo images and then a correspondence is established between them. In intensity based algorithms, blocks of pixels from one image are compared to blocks of pixels from other images and a similarity measure such as correlation or sum absolute difference is used to find the best matching block. The disparity value corresponding to the best matching block or patch is assigned to all the pixels of that template patch. Hence this is known as a patch based stereo correspondence algorithm. Each of these algorithms have their own advantages and disadvantages. Both algorithms are used widely.

The Intensity based algorithms are very simple to implement, they are robust and they produce dense depth maps. They fail to perform well when the distance between the stereo cameras is too large or if there are rotations and shears in the stereo images. However the major drawback of the intensity based algorithms is that they are highly computational and hence it will be the algorithm of interest in this paper. In this study we address the issue of high computational load of the intensity-based algorithms through novel modifications to the algorithm and by the way of analog signal processing.

## 2.1. Normalized Cross Correlation (NCC)

### 2.1.1. Review of Related Work

There has been a lot of research done on stereo image registration techniques as it relates to multiple fields like computer vision, medical imaging, photography etc. A variety of algorithms, both feature based and intensity based have been developed. In [1], the author provides a survey of different image registration techniques used in various fields.

In this study we are mainly concerned with the implementation of an intensity based image alignment algorithm in hardware. There has been some work done in this regard but most of them are improvements to the old algorithms and some are digital hardware implementations of these algorithms.

In [2] Lewis proposes a fast normalized cross correlation algorithm, which reduces the computational complexity of the normalized cross correlation algorithm through the use of sum table methods to pre-compute the normalizing denominator coefficients. In [3] the authors take the fast normalized cross correlation algorithm one step further by using rectangular basis functions to approximate the template image. The number of computations in the numerator will then be directly proportional to the number of basis functions used to represent the template image. Using a smaller number of basis functions to represent the template image will certainly reduce the computation but it may give a bad approximation of the template image, which would result in poor image alignment. In [4] the author uses a pipelined FPGA architecture to perform the Normalized Cross Correlation operation. This increases the computation speed significantly.

There have been various other improvements and implementations of the NCC algorithm in literature however none of the implementations, to our knowledge, try to tackle the computational intensity problem of the normalized cross correlation algorithm from an analog signal processing perspective.

### 2.1.2 Reasons for Choosing Normalized Cross Correlation

There are a lot of intensity based stereo correspondence algorithms. We chose Normalized Cross Correlation (NCC) as the algorithm that we would implement because of the following reasons:

1. The images being aligned in the patch based approach have translation in the X and Y direction but no rotation or shear. NCC algorithm performs well for such images.
2. NCC is less sensitive to variation in the intensity values of two images being aligned.
3. The NCC algorithm is computationally intensive. Hence it would be challenging to come up with methods to reduce the computation and make it implementable in real time or near real time and we believe analog signal processing would have a lot of value in such situations.

### 2.1.3 The General Algorithm

Template matching is one of the simplest methods used for image alignment. There are two images to be aligned. One image is called the template and the other image is called the reference. The template image is generally divided into blocks of smaller images. There is always a tradeoff between the depth accuracy that can be achieved and computation that can be handled in a NCC algorithm. Increasing the number of blocks by reducing the block size increases the accuracy to which depth can be estimated but it also increases the number of times the computations have to be performed. There is also a limit to which the block size can be reduced. If the block size is made too small then it might not have enough information to align with a matching block. Therefore choosing an optimum template block size is important.

Each template block is shifted on top of the reference image and at each point a correlation coefficient is calculated. This correlation coefficient will act as a similarity metric to identify the closest matching blocks.

The disadvantage of using cross correlation as a similarity measure is that it is an absolute value. Its value depends on the size of the template block. Also the cross correlation value of two exactly matching blocks may be less than the cross correlation value of a template block and a bright spot. The way around this problem is to normalize the cross correlation equation.

Equation (1) shows how the NCC algorithm is implemented [2].

$$C(u,v) = \frac{\Sigma_{x,y}\left[r(x,y) - \bar{r}_{u,v}\right]\left[t(x-u,y-v) - \bar{t}\right]}{\left\{\Sigma_{x,y}\left[r(x,y) - \bar{r}_{u,v}\right]^2 \Sigma_{x,y}\left[t(x-u,y-v) - \bar{t}\right]^2\right\}^{0.5}} \tag{1}$$

In the above equation $\bar{t}$ represents the mean of the template image block and $\bar{r}_{u,v}$ is the mean value of the reference image present under the template image block. The two summation terms in the denominator of the above equation represent the variances of the zero-mean reference image and template image respectively. Due to this normalization, the correlation coefficient is independent of changes to image brightness and contrast.

The denominator of the NCC equation can be calculated efficiently through the use of sum tables as suggested in [2]. However the numerator of the NCC is still computationally intensive. A direct implementation of the numerator of NCC algorithm on a template image of size ($T_x$ x $T_y$) and a reference image of size ($R_x$ x $R_y$) would require ($T_x$*$T_y$) multiplications and additions for each shift (u,v). Reducing the template image size would reduce the number of computations per block but it will also increase the total number of blocks on which the NCC has to be performed.

## 2.2. Modification to the NCC Algorithm

In a general normalized cross correlation algorithm the template image is divided into blocks and each block is shifted on top of the reference image. At each shift a normalized correlation coefficient is calculated. All the pixels in the block are used to perform this calculation as shown in equation (1). Once this is done for all shifts, a best matching block is picked and all the pixels in the template block are assigned the same shift/disparity value.

In the worst case scenario where there is no information available about the camera system or the scene, a brute force approach has to be used where the template image blocks have to be shifted all over the reference image. The computational complexity in this case would be very high. When some information is available about the camera system the maximum disparity that will be observed can be calculated and hence the number of shifts can be restricted. However this does not address the fact that the number of computations that have to be performed per block for each shift is still high.

A pre-processing step that is generally used in most stereo correspondence algorithms is image rectification. Image rectification projects stereo images onto a common reference plane so that the correspondence points have the same row coordinates. This essentially transforms the 2D stereo correspondence problem to 1D. However the rectification process itself will add to the computational complexity of the algorithm.

In order to further reduce the computation and address the above mentioned issues we decided to use only the diagonal elements of the template image block and the reference image blocks to compute the correlation coefficient. All the other steps in the algorithm are followed as given by equation (1). The thought behind this approach is that the diagonal elements of a block have enough information to calculate the disparity. By introducing this modification we have effectively converted the problem of 2D NCC operation to a 1D NCC operation. This is very similar to the image rectification operation but since we are choosing only the diagonal elements we introduce two advantages both of which contribute to the reduction in computation.

1. We are not using an algorithm to reduce the NCC operation from 2D to 1D, it is a natural result of the data selection process and hence it does not involve any additional computations.
2. Since we are choosing only the diagonal elements, the number of computations per block is reduced to a great extent i.e. if we have a template image of size $T_x$ x $T_y$ the total computations per block in the numerator now reduces from ($T_x$ x $T_y$) additions and multiplications to only $T_x(=T_y)$ additions and multiplications per shift. This reduction in the computation is even more significant when the template and reference image sizes are very large.

Another advantage of the modified NCC algorithm is in situations where the information of the camera systems capturing the images is not known and brute force shifts have to be applied. This modified algorithm will be a good solution for such cases.

In some pathological cases the images or particular template blocks might not have a lot of features like edges and contours or the features may all be located on the upper or lower triangular side. In such situations just using the diagonal elements to perform the NCC algorithm might not work. The solution to this problem may be as simple as using the off-diagonal elements instead of the diagonal elements. However in most practical applications we always have images with some features on the diagonal element so using all the pixels in a template image block is unnecessary as it will only add to the computation without producing any significant improvements in the results.

## 2.3. Hardware Architecture

In a standard CMOS image sensor there are photodiodes that produce electrons proportional to the amount of light intensity that strikes them. This is then converted into voltage levels which are read out by the readout circuitry. In order to remove noise, a process called correlated double sampling (CDS) is used. After this the signals are amplified. All this happens in the analog domain. These signals are then digitized using analog-to-digital converters (ADCs) and stored in memory or are sent to digital processors for further processing. So the signal for the most part is in the analog domain and we can utilize this to our advantage to perform analog processing.

With this in mind we have come up with a new imaging architecture which would best utilize the features of both analog and digital domains. Figure 1 shows a top level block diagram of the proposed architecture. In this architecture we have the digital system accessing the sensor and it is pre-processing, digitizing and storing the images in memory as before. However we now have an analog system that is accessing the analog data on the sensor, processing it and then feeding it into a digital machine for any further computations. By doing this we have separated the process of image acquisition which is being done by a digital system and image processing which is being done by an analog system. The biggest advantage of such an architecture is that they are operating in parallel i.e. the image acquisition is independent of the processing. This is not true in the case of completely digital systems. In a fully digital system the image processing operation cannot start until the images have been completely acquired and stored in memory. In this hybrid system the analog block is performing the computationally intensive task of running the NCC algorithm as and when the image data is being read off the sensor. This means that by the time the digital system has acquired the images the analog processor would have finished its computation and the outputs will be ready to be used by the digital system.

Another important point to be noted is that the analog processing block is not in the signal plane but in the control plane. One of the biggest disadvantages of an analog system is the amount of noise added by it. However in this kind of an architecture the analog block is not responsible for signal acquisition and hence the problem of signal being corrupted by noise vanishes immediately.

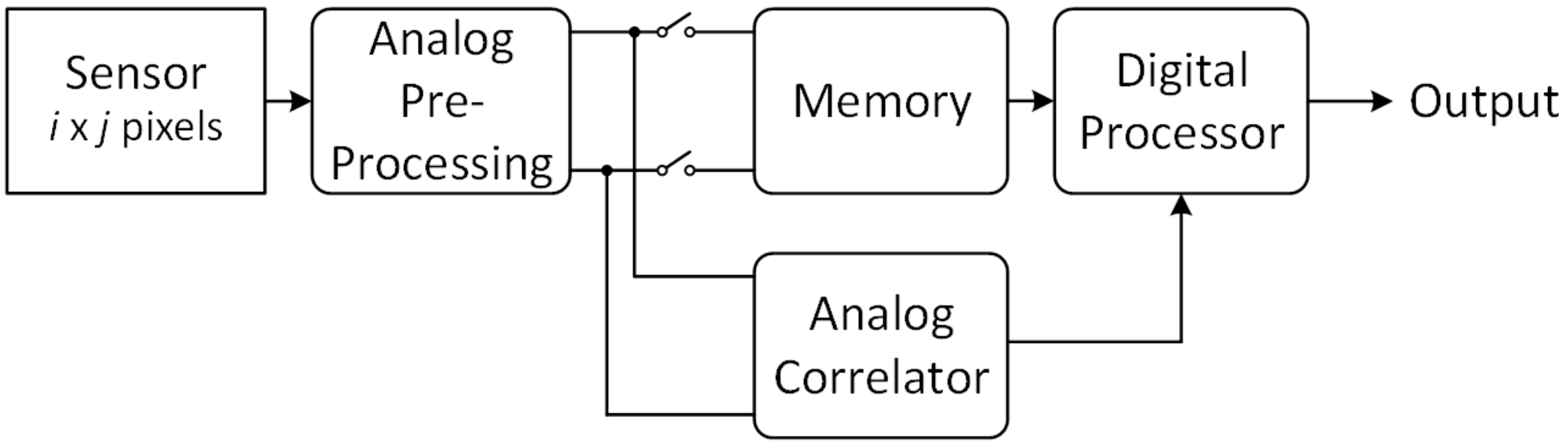

Figure 1: Proposed Architecture for the new Imaging systems

### 2.3.1. Implementing the modified NCC algorithm

Figure 2 shows the implementation of the modified NCC algorithm in the new imaging architecture. We have a digital system (shown on the left) which is accessing the sensor data, pre-processing, digitizing and storing it in memory. We have N analog channels which read the analog data from the sensor directly.

These N analog channels can be grouped into pairs in which one (odd numbered) channel is used to read the reference image data and the other (even numbered) channel is used to read template image data. The CMOS image sensors are capable of accessing individual pixel data. The readout circuitry is used to selectively read the diagonal elements of template and reference image blocks. Once the analog data has been read from the sensor we have it available to perform the NCC algorithm. According to eq. (1) in order to perform the NCC algorithm we first need a zero mean template image and zero mean reference image. In the case of digital systems it is not hard to compute the zero mean images from stored information. However in the new architecture we are directly accessing the analog data from the sensor and we would have to wait for an entire block of data to be read out in order to compute the mean and subtract it from the original signal. This would be a waste of processing time since this has to be done multiple times *i.e.* for each image block. In order to get around this problem we propose a second modification to the NCC algorithm. Here we use moving averages instead of regular averages and the moving average is subtracted from the original signal. The moving average circuit can be implemented as a low pass filter. The analog data from the sensor is passed through the moving average filter, the output of which is subtracted from the original signal. This is then fed to the multiplier and integrator which together perform the correlation operation. This calculates the numerator of the NCC algorithm. For calculating the denominator we plan to use the sum table method. This can be done efficiently by a digital system. So once the numerator calculation is done the analog signal is sampled and then fed into a DSP which normalizes the numerator and performs the decision making. The output will be disparity values in the X and Y direction. We also ran MATLAB simulations to ensure that a change from an actual zero-mean image to one with moving averages does not affect the performance of the NCC algorithm.

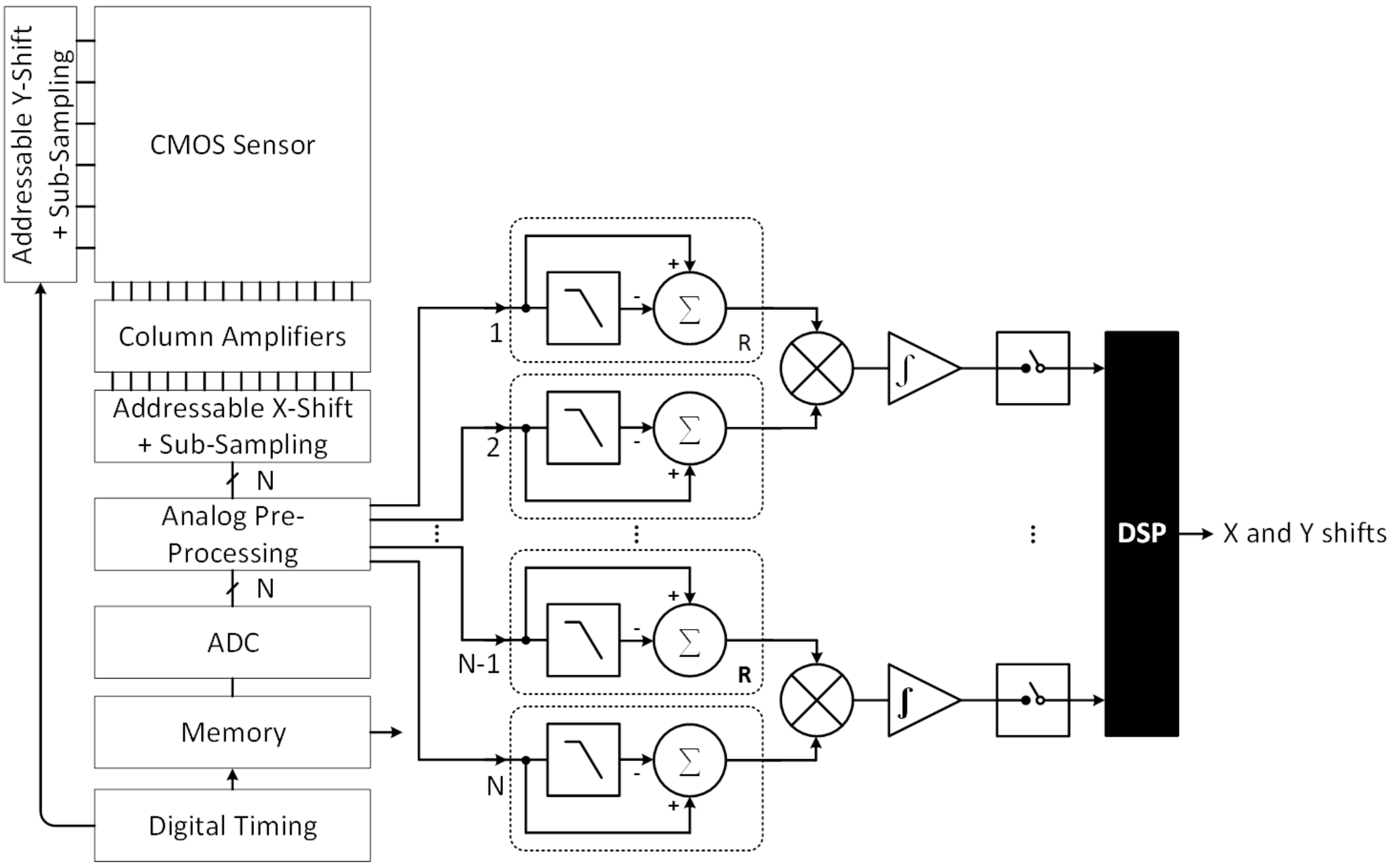

Figure 2: Implementation of NCC algorithm in the new Imaging architecture

## 2.4. Experimental Results

In this section we compare the performance of the modified NCC algorithm to the original algorithm to show that the modified algorithm is faster and has a performance similar to the original algorithm. Since the modified algorithm has been developed to be implemented in analog hardware various other simulations are run that measures the performance of the modified algorithm.

We have run simulations on 15 sets of unrectified, grayscale stereo image pairs. These images have been captured under different illumination conditions which include incandescent light (In_Incd), outside bright light (Out_brt), outside low light (Out_clds), outside mixed shade lighting (Out_mxdshd). This allows us to test the performance of the algorithm in a more robust manner. All simulations have been done in MATLAB.

### 2.4.1. Cropping the template image

The stereo cameras have overlapping fields of view but the amount of overlap depends on the distance between the centres of the two cameras. In this work we have considered cameras whose centres are 6.5mm apart. Their hyperfocal distances are 70cm which means that all objects which are 35cm and beyond are in sharp focus. Since the distance between the centres are 6.5mm there will be some points in the scene that will be present in one of the images but not in the other. These points cannot be used for alignment and hence one of the images (template image) is cropped around the edges.

### 2.4.2. Evaluating the performance of the algorithms

The performance of the image alignment algorithms can be evaluated in a variety of different ways. Here the correlation coefficient is used as a performance measure. Once the final disparity values for all the template blocks are obtained, each template block is shifted by the disparity values obtained for that block. In order to get a uniform disparity variation across the entire image an interpolation technique is used. At the end of this process the two stereo images have been aligned. The correlation coefficient is calculated between the two aligned images and it is used as an indicator of the performance of the algorithm.

Figure 3 shows a comparison of the correlation coefficients obtained for 15 stereo image pairs by using the original NCC algorithm and the modified NCC algorithm. Each stereo image pair has a size of 1080x1920 pixels. The cropping of the template image around the edges is not more than 10% of the entire image size. The template block size chosen here is 128x128 pixels. As can be seen from the figure the performance of the modified NCC algorithm is very close to the original NCC algorithm. The performance was also tested for various other template block sizes and uniform performance was obtained for all. A speedup of 2x in MATLAB run time was observed for the modified NCC algorithm over the original algorithm.

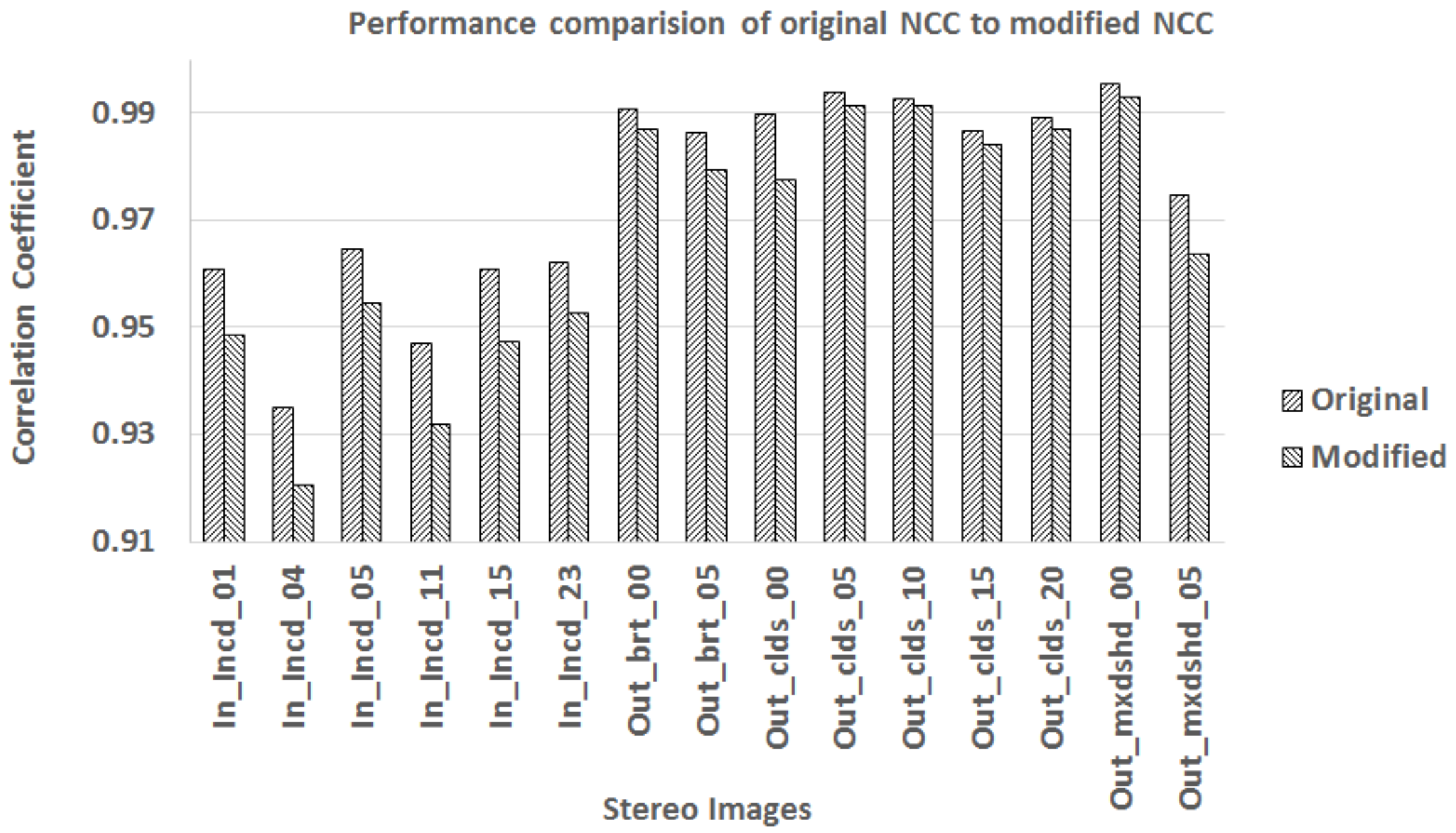

Figure 3: Performance comparison of the original NCC algorithm to modified NCC algorithm

It has to be noted that the objective of the above simulation is to compare the performance of the algorithms and not the digital and analog implementation of the algorithms.

As an example of the performance the modified NCC algorithm two figures 4 and 5 are shown. Figure 4 shows an overlap of a pair of stereo images before alignment. The areas of magenta and green show the areas of misalignment between the two images. The correlation coefficient measured for these two images before alignment is 0.7247. Figure 5 shows the overlap of two images after aligning them using the modified NCC algorithm. As can be seen from the figure there is hardly any misalignment between the two images.  The correlation coefficient observed in this case is 0.9923.

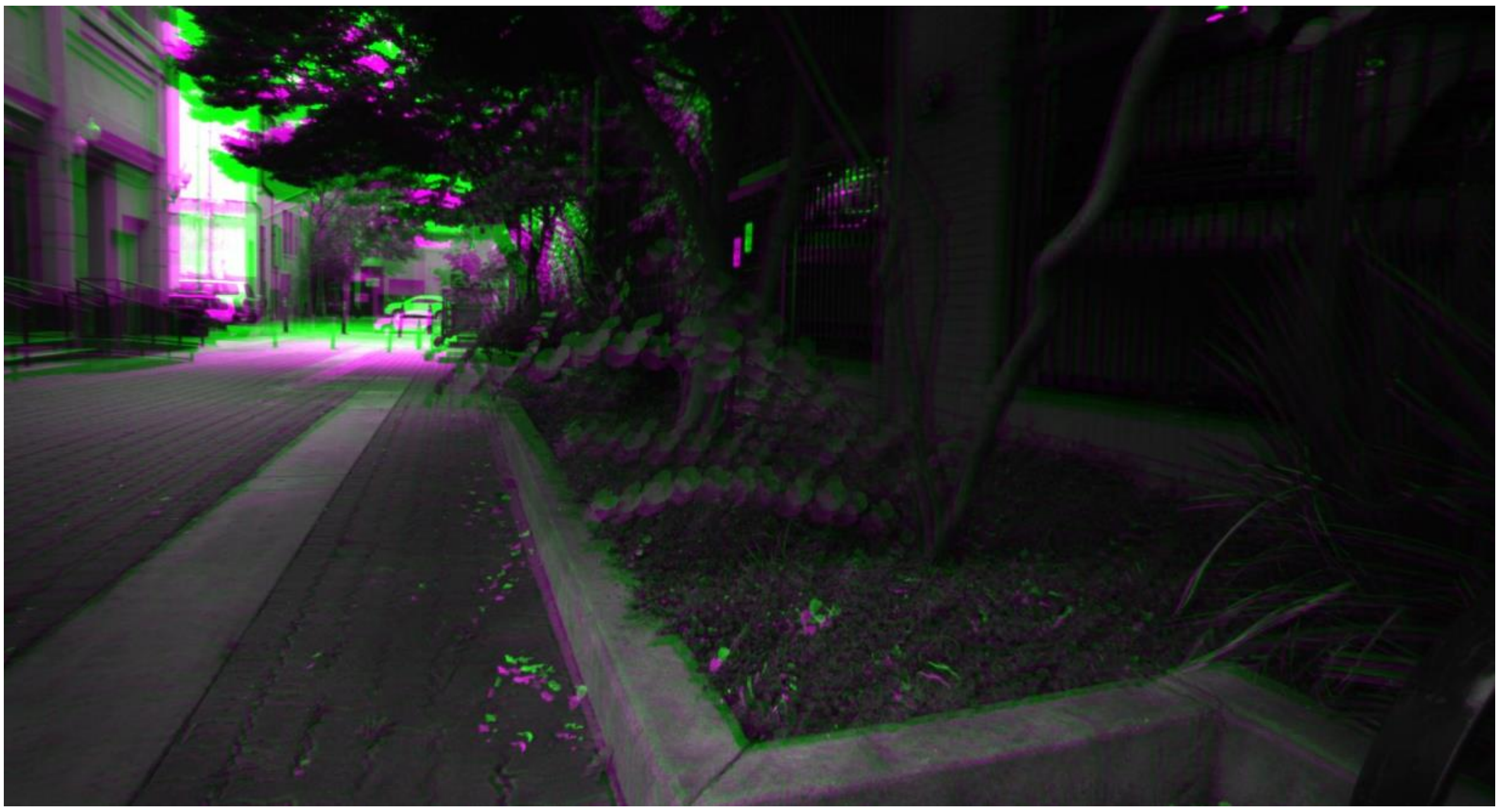

Figure 4: Overlap of two stereo images before alignment

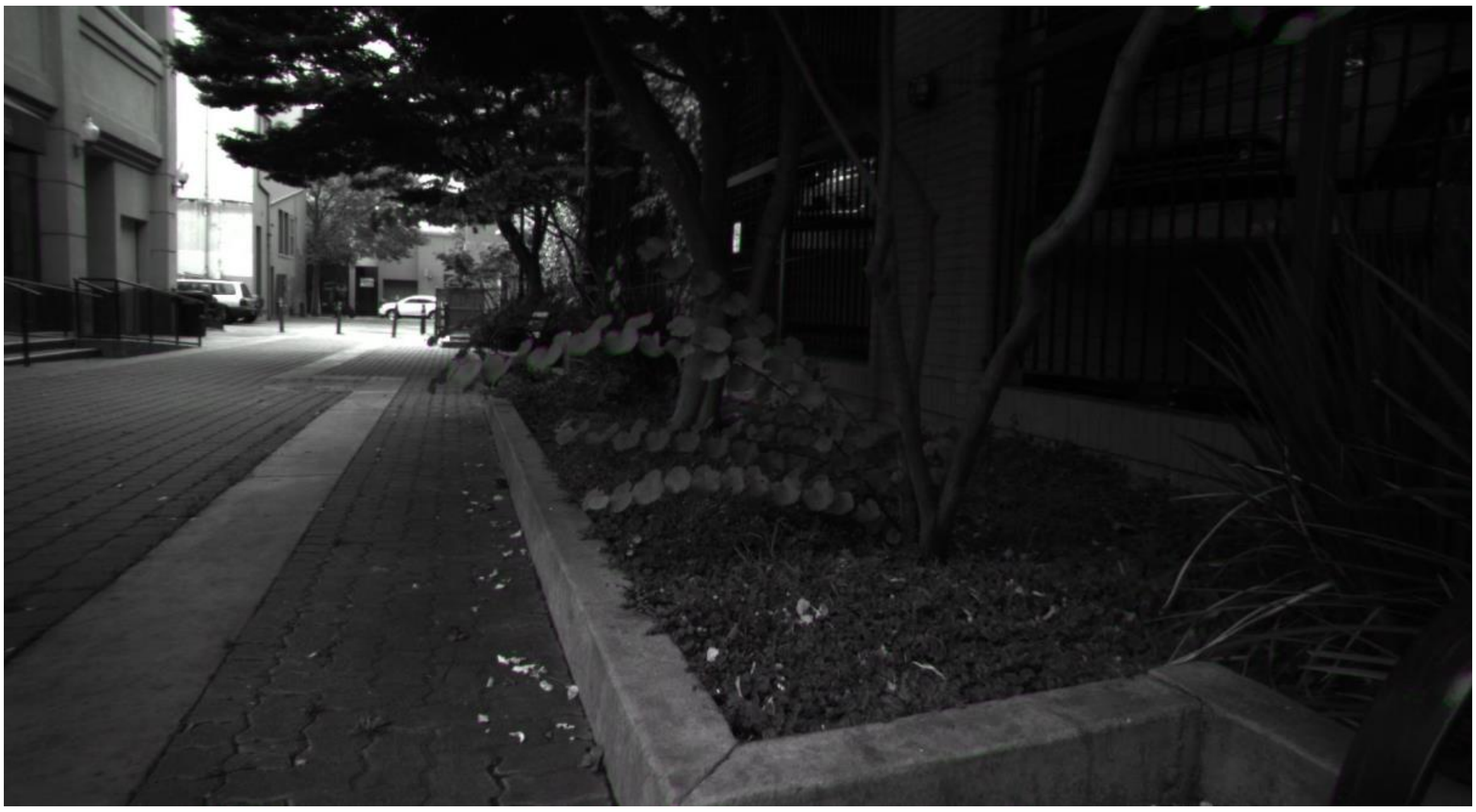

Figure 5: Overlap of the two stereo images after alignment

Figure 6 and Figure 7 shows the disparity variation for the image shown in figure 4 and 5 in X and Y direction respectively. These disparity values have been obtained through the modified NCC algorithm. The disparity values have been colour coded and the colour bar indicates the different disparity values. The disparity values vary from 18 to 33 for Figure 6 in the X (horizontal) direction. The disparity values vary from 43 to 53 for Figure 7 in the Y (vertical) direction.

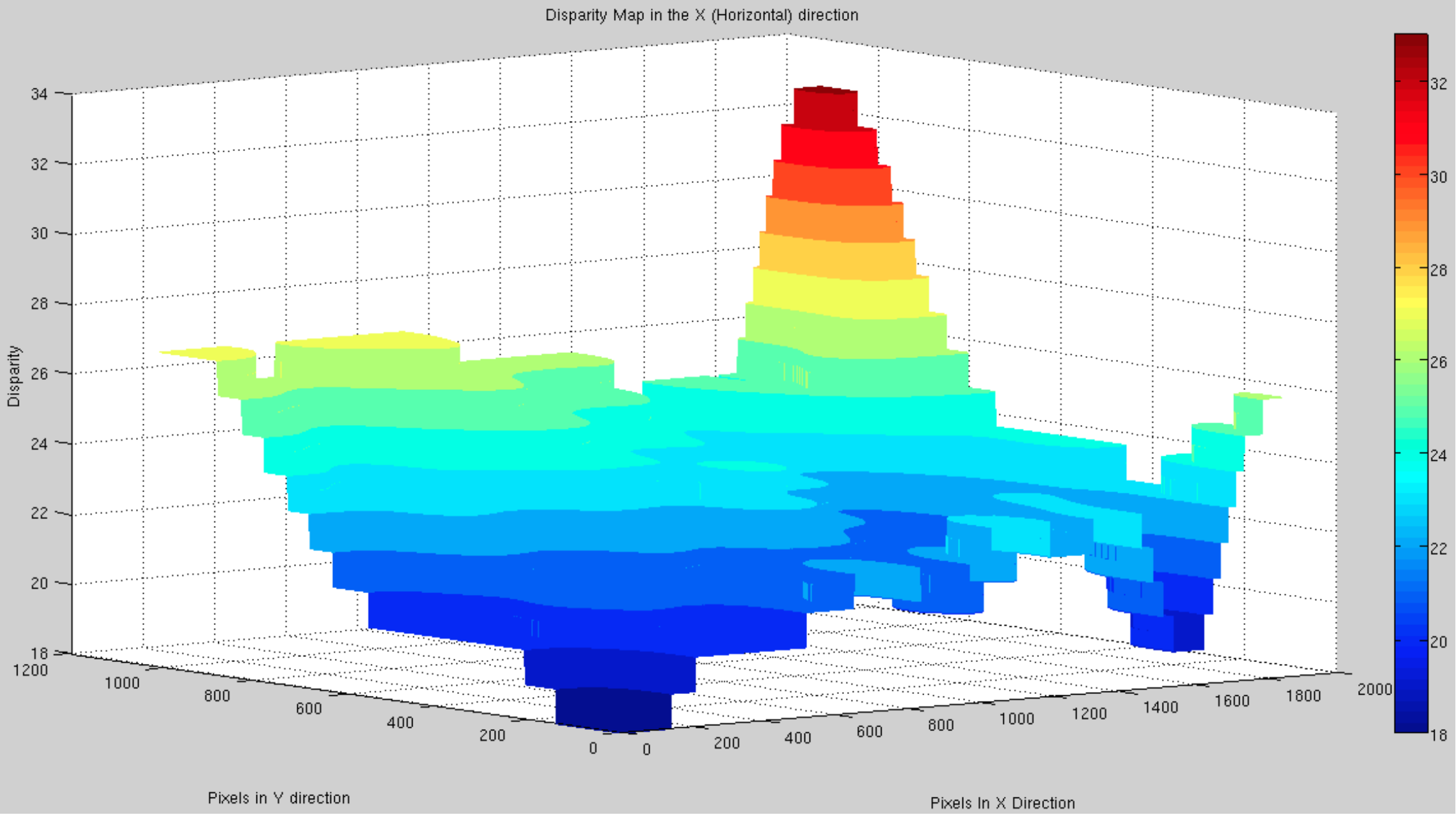

Figure 6: Disparity variation in the X(Horizontal) direction

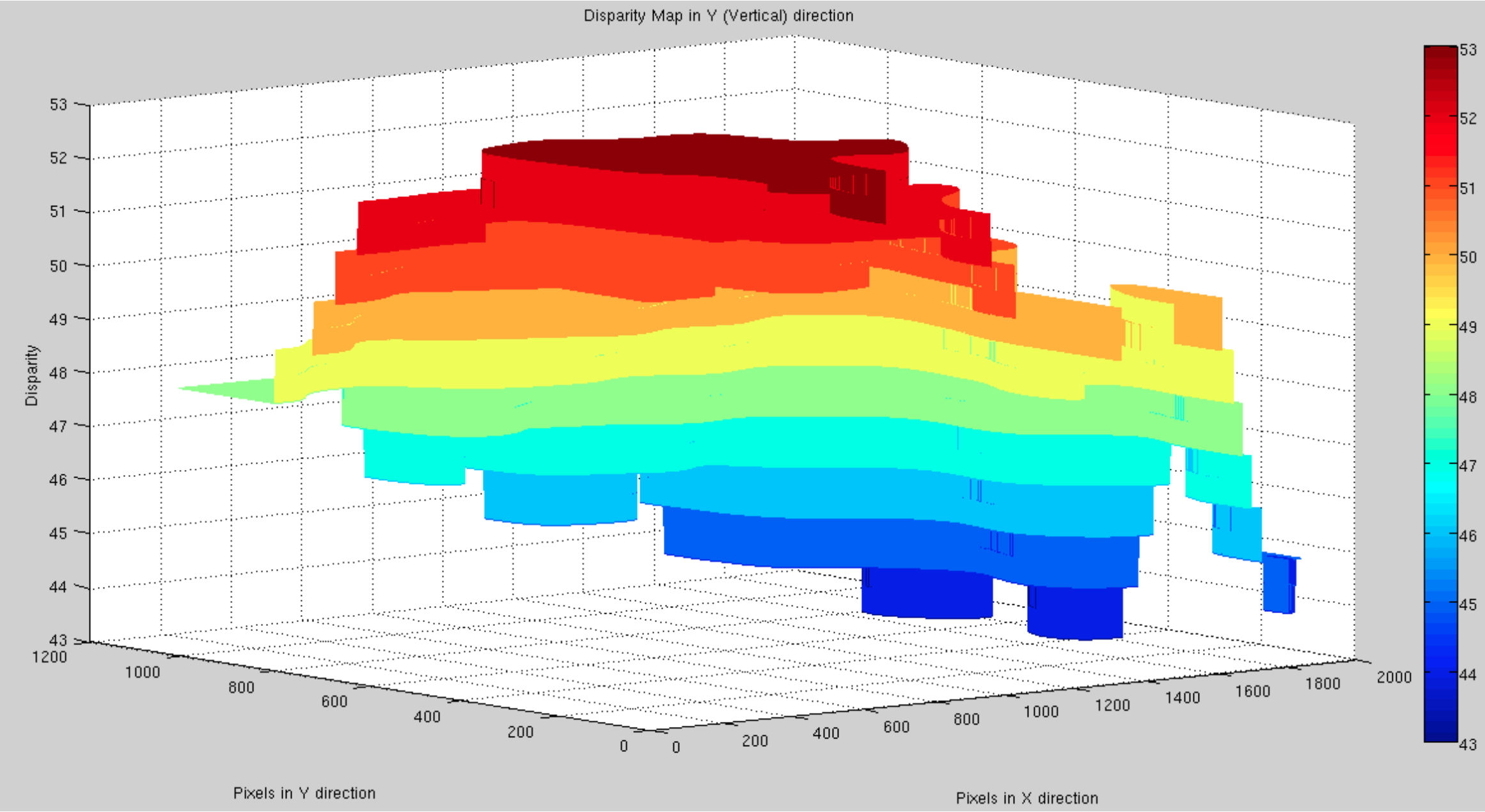

Figure 7: Disparity variation in the Y(Vertical) direction

### 2.4.3 Measuring the robustness of the modified NCC algorithm

We know that the NCC algorithm is robust to changes in the intensity values of the images. Here we try to measure the robustness of the modified NCC algorithm to changes in intensity values by changing the intensity values of one of the two stereo images. In the first case we reduced the intensity values of the template image by 90% uniformly across the image and used the modified NCC algorithm to align the images. This analysis addressed the fact that the illumination of a scene might change between the capture of two stereo images and the change in illumination was assumed to be uniform. However there might be rare situations where illumination on parts of the scene varies between captures of two stereo images. To address this issue we randomly varied the intensity values of the template image and analysed the performance of the modified NCC algorithm under these conditions as well. Figure 8 shows the results of these analyses. As it can be seen the modified NCC algorithm is very robust to changes in the image intensity values.

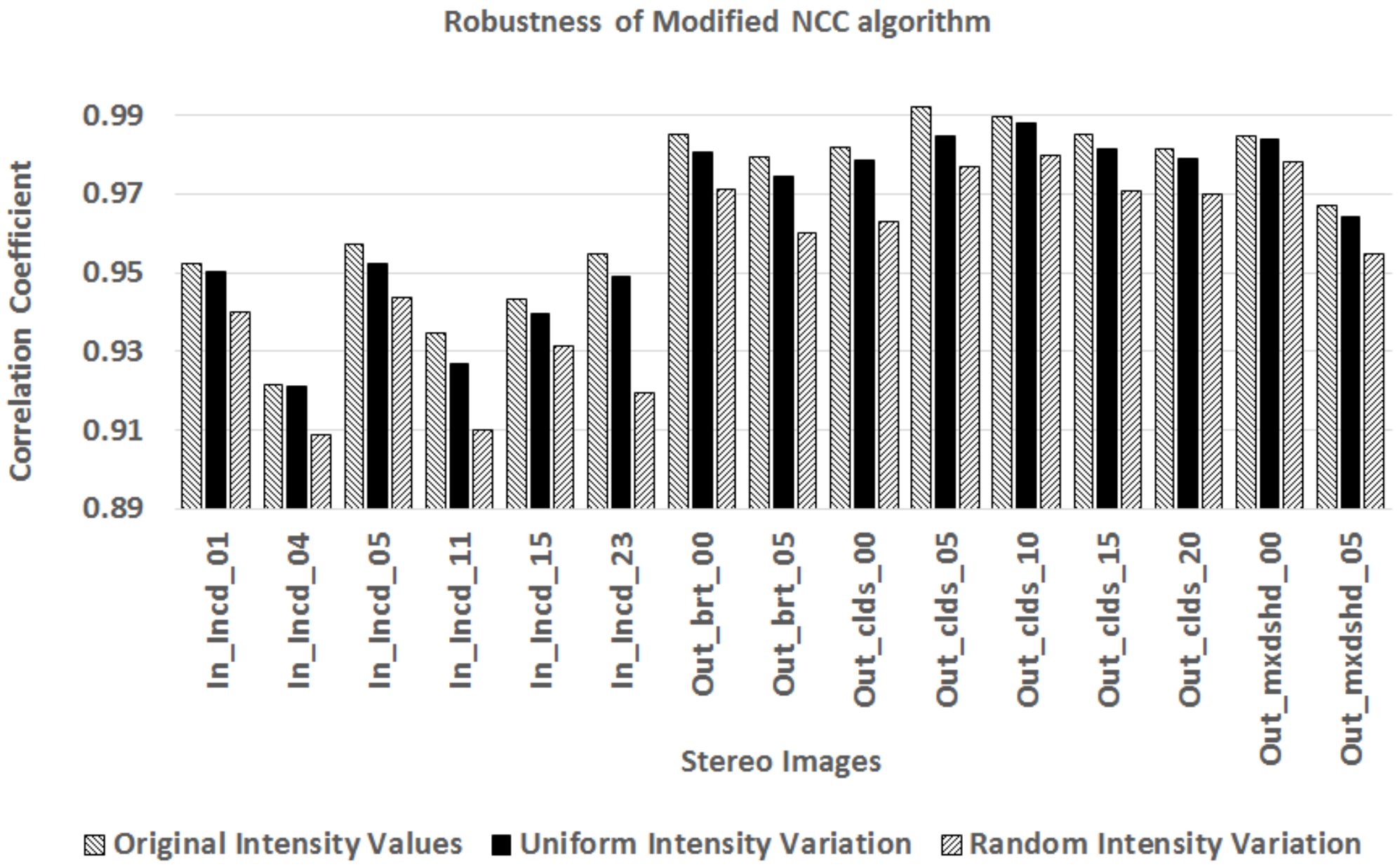

Figure 8: Robustness of the modified NCC algorithm to changes in image intensity values

Since the algorithm has been developed to be implemented in analog hardware it is very important to characterize the performance of the algorithm in the presence of noise. The two analog circuits that have been considered to be the primary contributors of noise are the multiplier and the integrator. In order to simulate the addition of noise by analog circuitry we first find the RMS value of the image intensity values. We multiply this RMS value by a number which indicates the percentage of noise being added by the circuit. This value is then multiplied by a random number picked from a Gaussian distribution. The outcome of this process is a noise value which is then added to the original image intensity. In our simulations it was found that the algorithm is more sensitive to the noise added by the multiplier than that by the integrator. Hence we maintain the noise added by the integrator at 20% and vary the amount of noise added by the multiplier. Figure 9 shows this performance variation. We have shown the performance for 3 different noise values added by the multiplier, 1%, 10% and 20%. As it can be seen the performance of the modified NCC algorithm is still good in the presence of noise added by the analog circuitry.

### 2.4.4. Dynamic Range requirements and Power analysis

The amount of noise that can be tolerated by a circuit determines the dynamic range requirements for that circuit. The noise analyses done above will give us an idea of the dynamic range requirements for the analog circuits to perform the NCC operation. Analog circuits can easily achieve dynamic ranges of 40dB. This corresponds to a noise level of 1%. From Figure 9 we see that the modified NCC algorithm has excellent performance for a noise level of 1%.

The dynamic range requirements also dictates the power consumptions for analog circuits. The 4 major analog circuits required to implement the modified NCC operation are the low pass filter, analog summer, multiplier and integrator. We have performed initial circuit simulations for these components. Table 1 shown below gives approximate values of the power consumption for these components calculated for a dynamic range of 40dB based on these simulations. Here we assume that we have 64 analog channels in the architecture shown in Figure 2 and the number of components required are calculated based on that. The actual power consumptions can only be obtained once the components are realized and integrated.

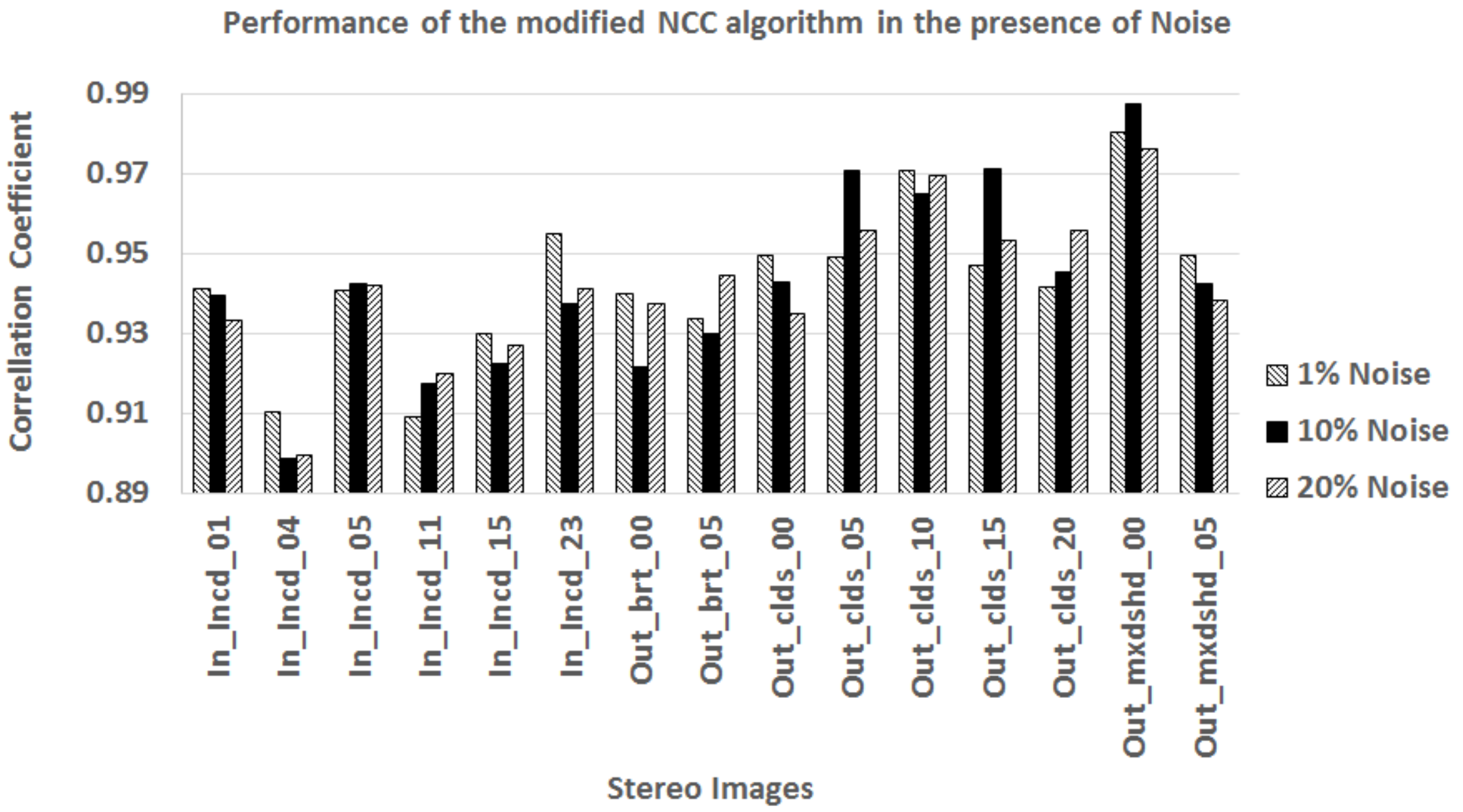

Figure 9: Performance of the modified NCC algorithm in the presence of Noise

Table 1: Approximate power consumption values for the analog circuits

| ***Component*** | ***Quantity*** | ***Power Consumption*** |
|---|---|---|
| LPF | 64 | 2.8mW/LPF*64 = 179.2mW |
| Summer | 64 | 0.549mW/sum*64=35.13mW |
| Multiplier | 32 | 1.83μW/mul*32 = 0.058mW |
| Integrators | 32 | 0.024mW/int*32 = 0.768mW |

The total power consumption by the analog circuitry is 215.15mW. Based on some of the digital implementations such as [11] and [12], we see that the power consumption for an analog implementation will be very low compared to that of a digital implementation. This shows that we have significant power savings as well.

### 2.4.5. A Note on Computation Time

The modifications proposed to the NCC algorithm contribute to a significant reduction in the computation of the algorithm. Simulation results show a 50% reduction in computation time for the modified NCC algorithm over the original algorithm. The other factors which add to the reduction of computation time are the novel imaging architecture and analog processing. In the new imaging architecture the analog processor works in parallel with the digital acquisition system and hence it does not have to wait for the entire image to be acquired before the processing starts. By the time the acquisition is done the analog processor would have finished its computation. So the image acquisition time can also be added towards the reduction in computation time. The implementation of the NCC algorithm is being done in analog hardware. The analog processor is not limited by the data converters (ADCs) or logic delays. The settling times of well-designed analog circuits are very small. Hence an analog implementation of the NCC algorithm would be faster than a digital implementation and would contribute towards a further reduction in computation time.

# 3. PER-PIXEL STEREO CORRESPONDENCE ALGORITHM

The patch based stereo correspondence algorithm discussed in the previous section produces very coarse depth maps. This approach is suitable for applications where speed is more important than accuracy. However in some of the other applications, a more accurate or a finer depth map is required. In these cases, a per-pixel approach of finding depth is preferred to a patch based approach. In this section we discuss a per-pixel stereo correspondence algorithm. The similarity measure used here is Sum Absolute Difference (SAD). We show that using a slightly lesser overlap between blocks can give a similar performance while significantly reducing the computational load. A hardware architecture is proposed which can further speed up computation.

## 3.1. Review of Related Work

There are some major challenges associated with recovering an accurate depth map like the presence of occluded pixels, noise in images and depth discontinuities. A multitude of stereo correspondence algorithms have been developed which address some or all of these challenges. In [14] the authors provide a thorough analysis and overview of the different stereo correspondence algorithms. One of the algorithms that has a good performance is proposed by Klaus et al. [15]. They use a segment-based stereo correspondence algorithm which uses belief propagation and a self-adapting dissimilarity measure.

## 3.2. Disparity Estimation Using Overlapping Blocks

Most of these per-pixel stereo correspondence algorithms calculate a coarse depth estimate of the scene using local window based approaches as the first step. One of the major problems associated with using the patch based approach for calculating the coarse depth estimates is that the algorithm does not perform well when there are depth discontinuities in a block. Also depending on the application the estimated depth may be too coarse.

In order to obtain a finer depth estimate and reduce the effect of depth discontinuities overlapping blocks can be used. The higher the overlap, lesser is the effect of depth discontinuities and a finer depth map is produced. If the blocks are completely overlapping (except for one pixel) then this method can be thought of as per-pixel depth estimation. The drawback of using such an approach is that computation increases with the amount of overlap. The computational intensity of such an algorithm is orders of magnitude higher than the patch based approach and the digital system will be bogged down by the high computational load. In order to get around this problem we can use analog signal processing and a hybrid architecture very similar to the one shown in Figure 1, the change being the similarity measure used.
The similarity measure being used here is Sum Absolute Differences (SAD) as shown in the equation below.

$$SAD(u,v) = \sum_{n=1}^{blkHeight} \sum_{m=1}^{blkWidth} |R(u,v) - T(u+n, v+m)| \quad \textbf{(2)}$$

In equation (2), R is the reference image block and T is the template image block. The parallel analog correlator in the architecture of Figure 1 will be replaced by an analog processor that can perform the above SAD operation.

Most of the stereo correspondence algorithms in the digital domain operate on rectified images. In the new architecture we have access to raw analog data directly from the sensor and not the rectified images. Hence in this implementation we check for disparity in both the x and y directions.

## 3.3. Hardware Architecture

Figure 10 below shows the implementation of the SAD algorithm in the new hardware architecture. As before we have N analog channels of which the odd numbered channels are reading the reference image data and the even numbered channels are reading the template image data. In order to have a faster computation the entire rows or columns of the reference and template image data are read at once and their difference is calculated. This data is then passed through an absolute value circuit and then through the integrator. The output of the integrator is sampled at a particular rate depending on the size of the window/block. The integrator accomplishes a 1D summation. These values are then fed into a DSP where the second dimensional summation is performed along with any further processing and decision making that is required. As before the image acquisition is independent of processing and they are happening in parallel. This leads to a significant reduction in computation time.

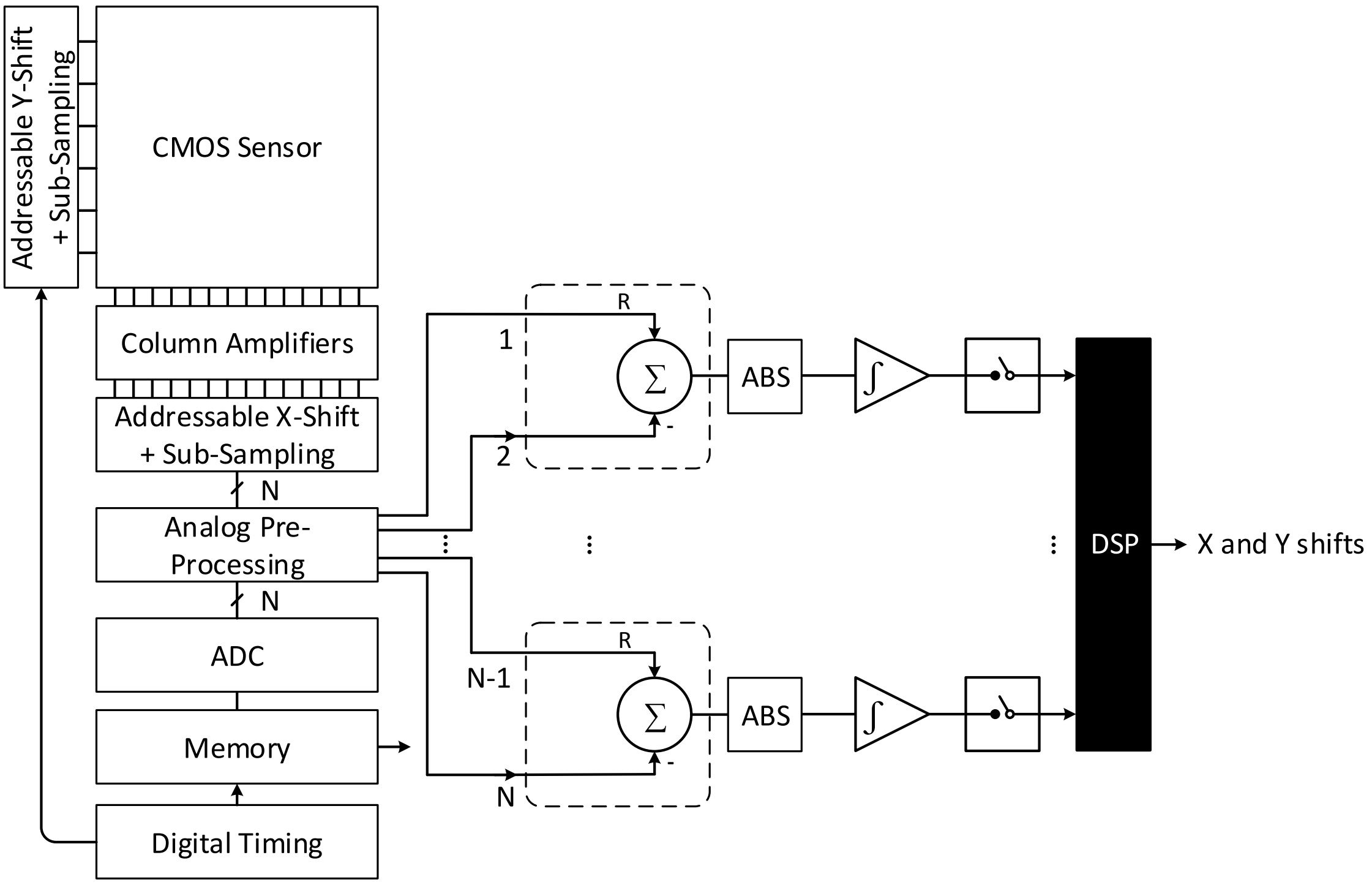

Figure 10: Implementation of the SAD algorithm in the new imaging architecture

Another feature that can obtained from this new architecture is the varying overlap of blocks. The amount of overlap determines the accuracy to which the depth or disparity can be obtained.

## 3.4. Experimental Results

In this section we test the performance of the SAD algorithm by performing various simulations in Matlab. Since the SAD algorithm will be implemented in analog hardware we have considered the performance of the algorithm in the presence of added noise by the analog circuitry. We have run simulations on two stereo images from the standard Middlebury dataset [16].

The performance of the image alignment algorithms can be evaluated in a variety of different ways. Here the performance measures used are correlation coefficient and RMS disparity error. Once the final disparity values for all the template blocks or pixels are obtained, each template block or pixel is shifted by the disparity values obtained. At the end of this process the two stereo images have been aligned. The correlation coefficient is calculated between the two aligned images and it is used as an indicator of the performance of the algorithm. The Middlebury dataset also has a ground truth disparity map available. This is used to compute the RMS disparity error between the ground truth disparity map and the computed disparity map which can also be used as a performance measure as suggested in [14].

Figures 11 and 12 shows the correlation coefficients obtained for the Middlebury dataset ‘Baby1’ and ‘Bowling1’ respectively. We have measured the correlation coefficient for 4 different block sizes 5x5, 7x7, 11x11 and 15x15 pixels. The overlap between the blocks is varied for each block size, from 0 pixels which represents a patch based approach to n-1 pixels which represents the per-pixel approach, where n is the size of the block. The x-axis in figures 11 and 12 represents the pixel overlaps. For each block size and overlap we compare the performance of the SAD algorithm with and without noise being added by the analog circuitry. The amount of noise added by the analog circuits is fixed at 5% of the RMS signal level. The procedure for adding noise is the same as explained in section 2.4.3. As can be seen from the figures the performance of the algorithm is good even in the presence of noise.

Figures 13 and 14 shows the comparison of the RMS disparity error between the ground truth disparity map and the computed disparity map for the Middlebury dataset ‘Baby1’ and ‘Bowling1’ respectively for 4 different block sizes. The overlap between blocks is also varied as before which is represented by the x-axis of figures 13 and 14. We compare the performance of the SAD algorithm both with and without the noise added by the analog circuitry. The amount of noise added is 5% of the RMS signal level. As can be seen from the figures the performance of the algorithm after the addition of noise is very close to the ideal case.

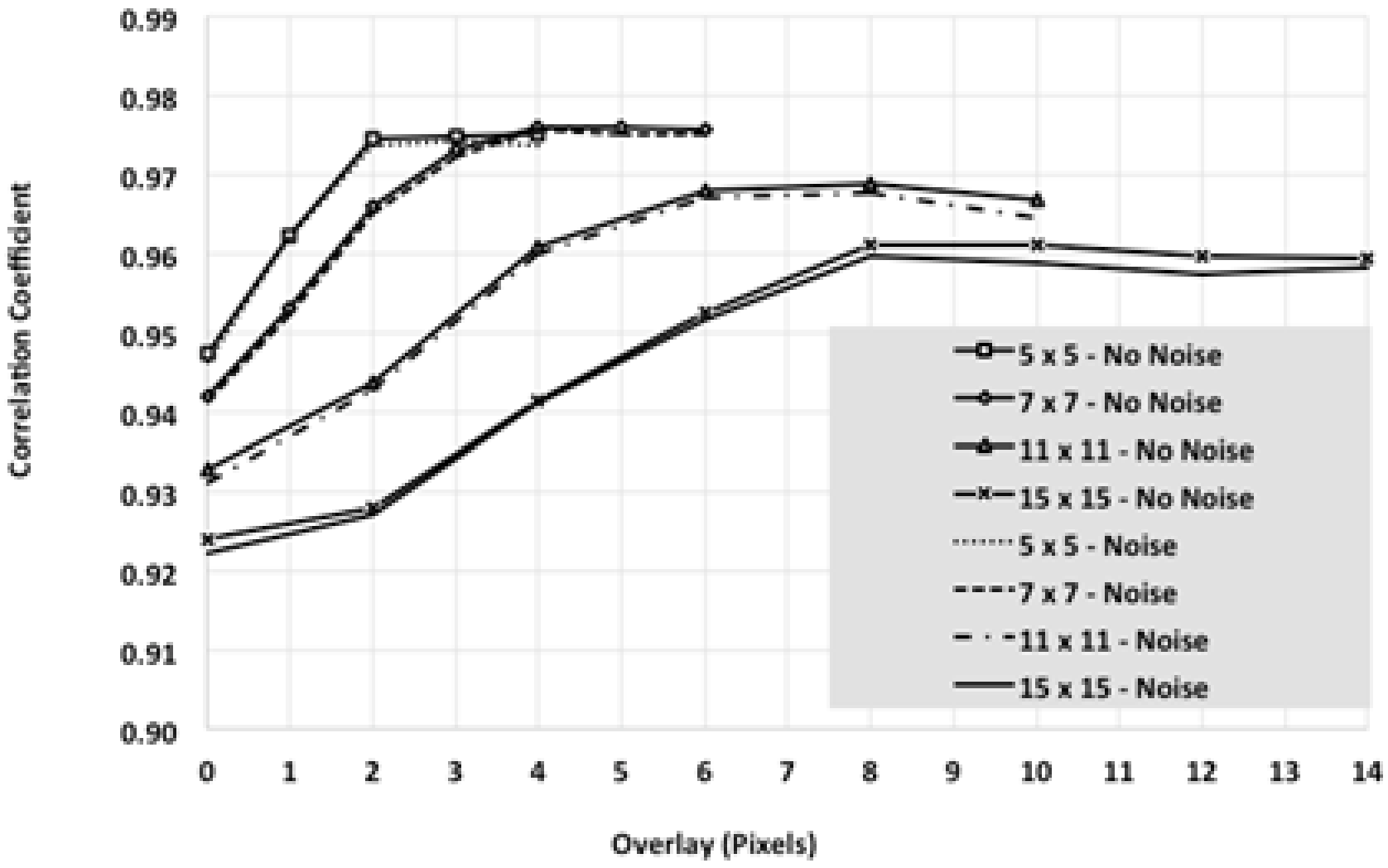

Figure 11: Correlation Coefficient for different block sizes with and without noise (Dataset: ‘Baby1’)

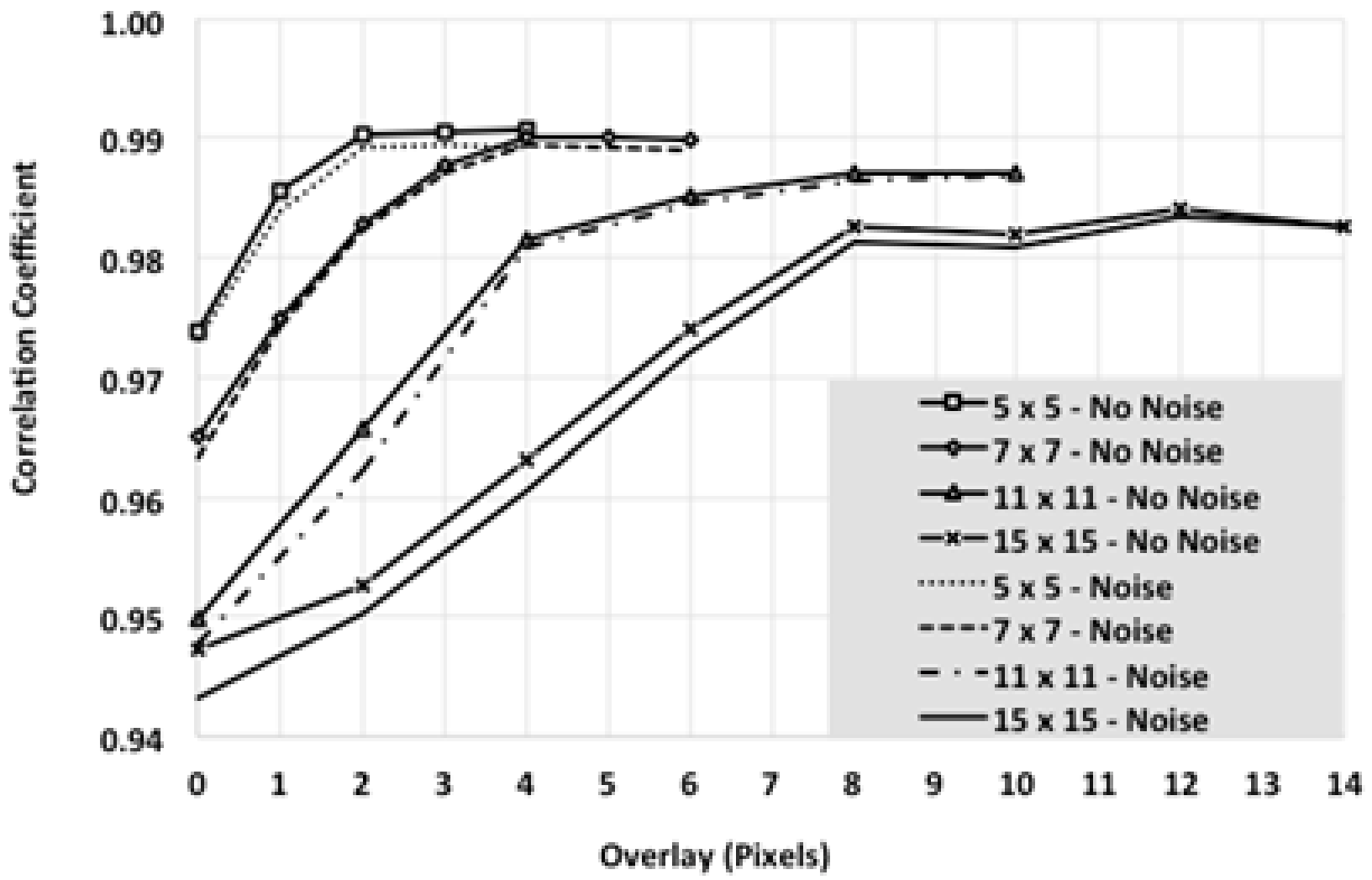

Figure 12: Correlation Coefficient for different block sizes with and without noise (Dataset: 'Bowling1')

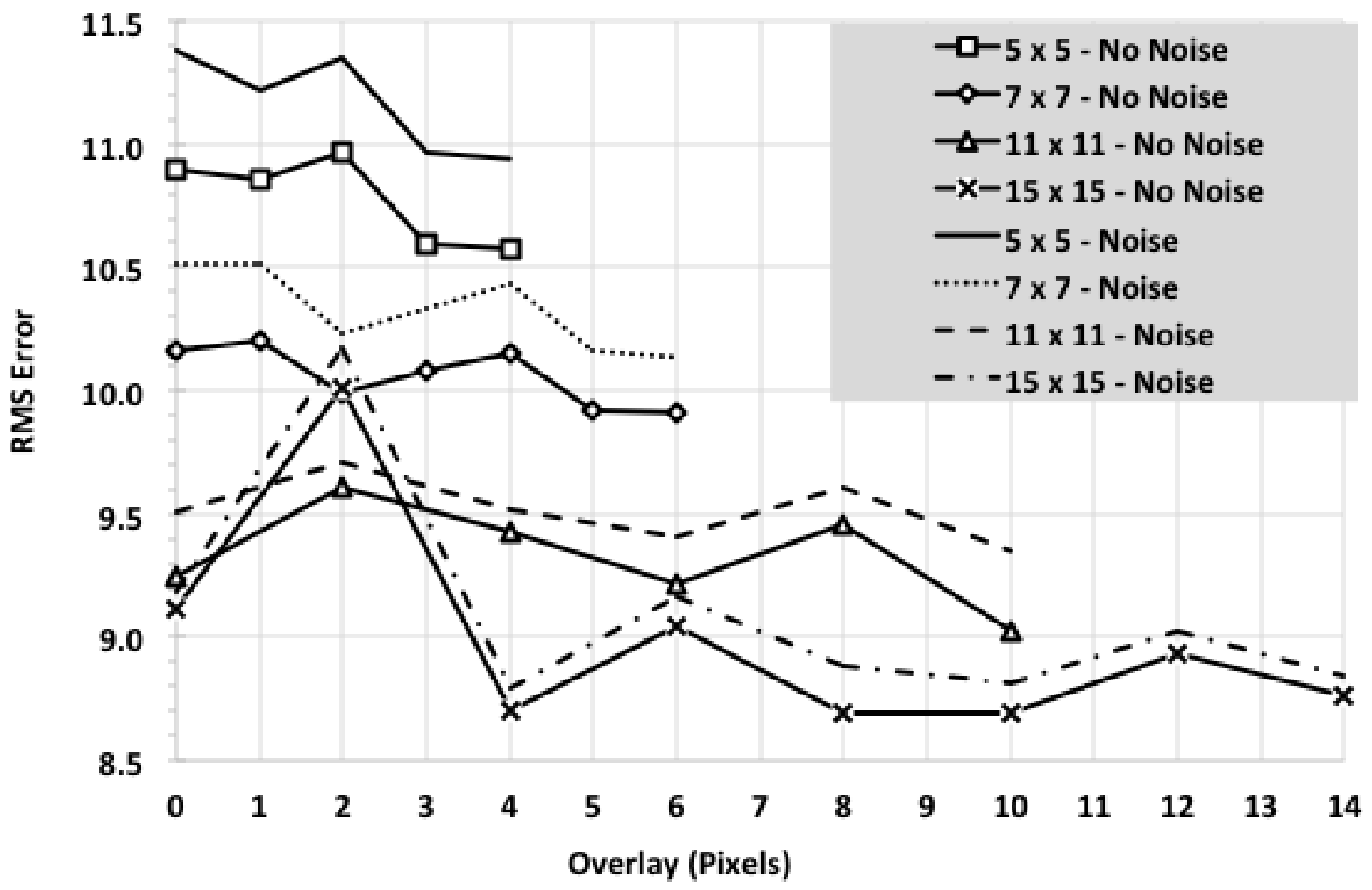

Figure 13: RMS disparity error for different block sizes with and without noise (Dataset: Baby1)

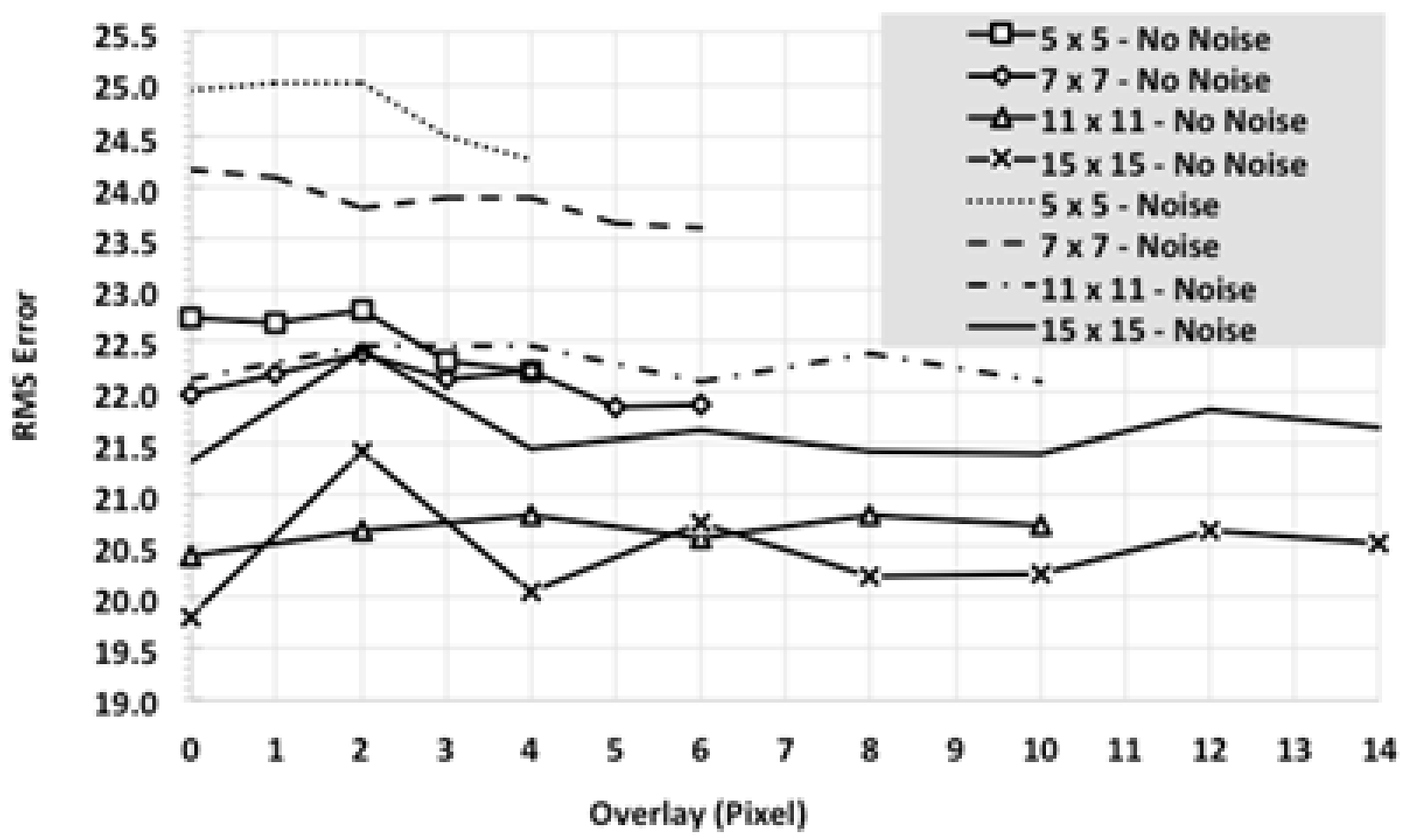

Figure 14: RMS disparity error for different block sizes with and without noise (Dataset: 'Bowling1')

Figures 15 shows the overlap of two stereo image pairs before alignment (left image) and after alignment (right image). Before alignment the correlation coefficient is 0.5189 and after alignment the correlation coefficient is 0.9906.

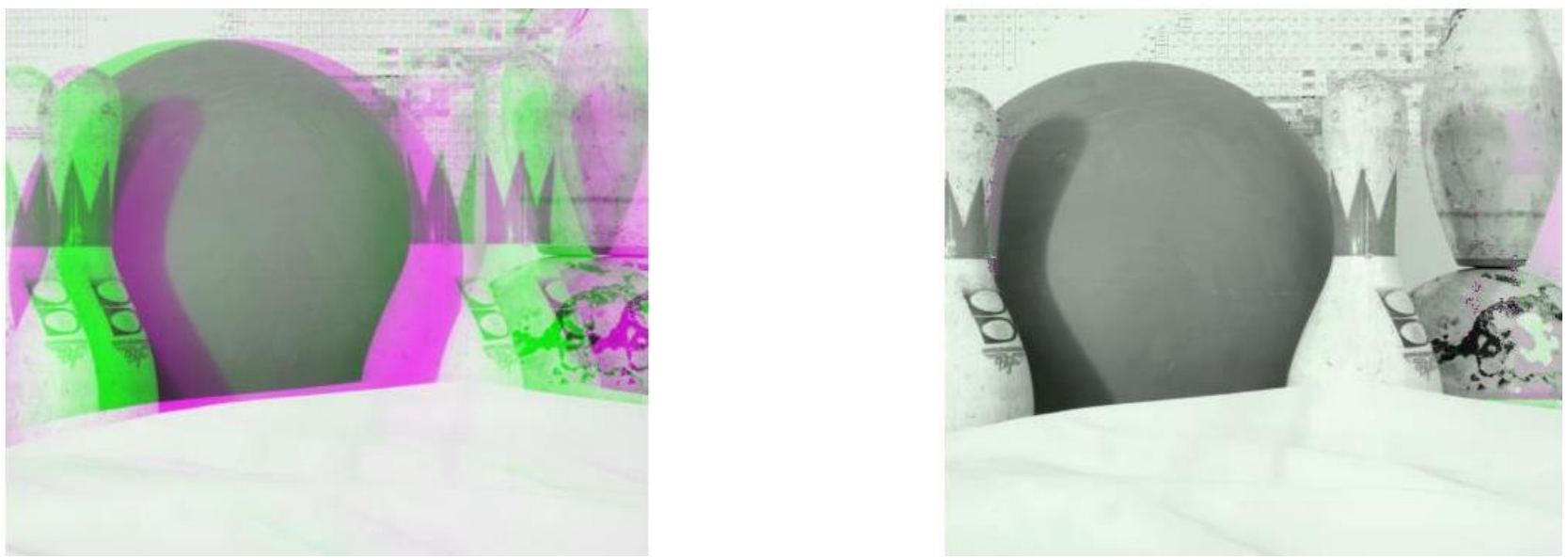

Figure 15: Overlap of stereo images before alignment (left) and after alignment (right) (Dataset: 'Bowling1')

Figure 16 shows the x disparity map obtained for the 'Bowling1' dataset for a window size of 5x5 pixels for different overlaps. The first row has 3 disparity maps with 0, 1 and 2 pixel overlap between blocks. The second row has 2 disparity maps with 3 and 4 pixel overlap between blocks. The disparity map has been generated after considering a 5% noise added by the analog circuitry during processing. As can be seen from the disparity map, as the overlap increases the disparity map generated gets finer but it hits a saturation level at a certain point after which we do not see a lot of improvement. This is also evident from the correlation coefficient curves in figures 11 and 12. This shows that we have to select an overlap factor which gives a good performance while not introducing unnecessary computation. We believe an overlap factor of 60% to 70% would be a good trade-off between computation and accuracy of the disparity map. By reducing the overlap the amount of computation is also reduced significantly. Due to this modification an average speed up of 4x was observed in Matlab run time.

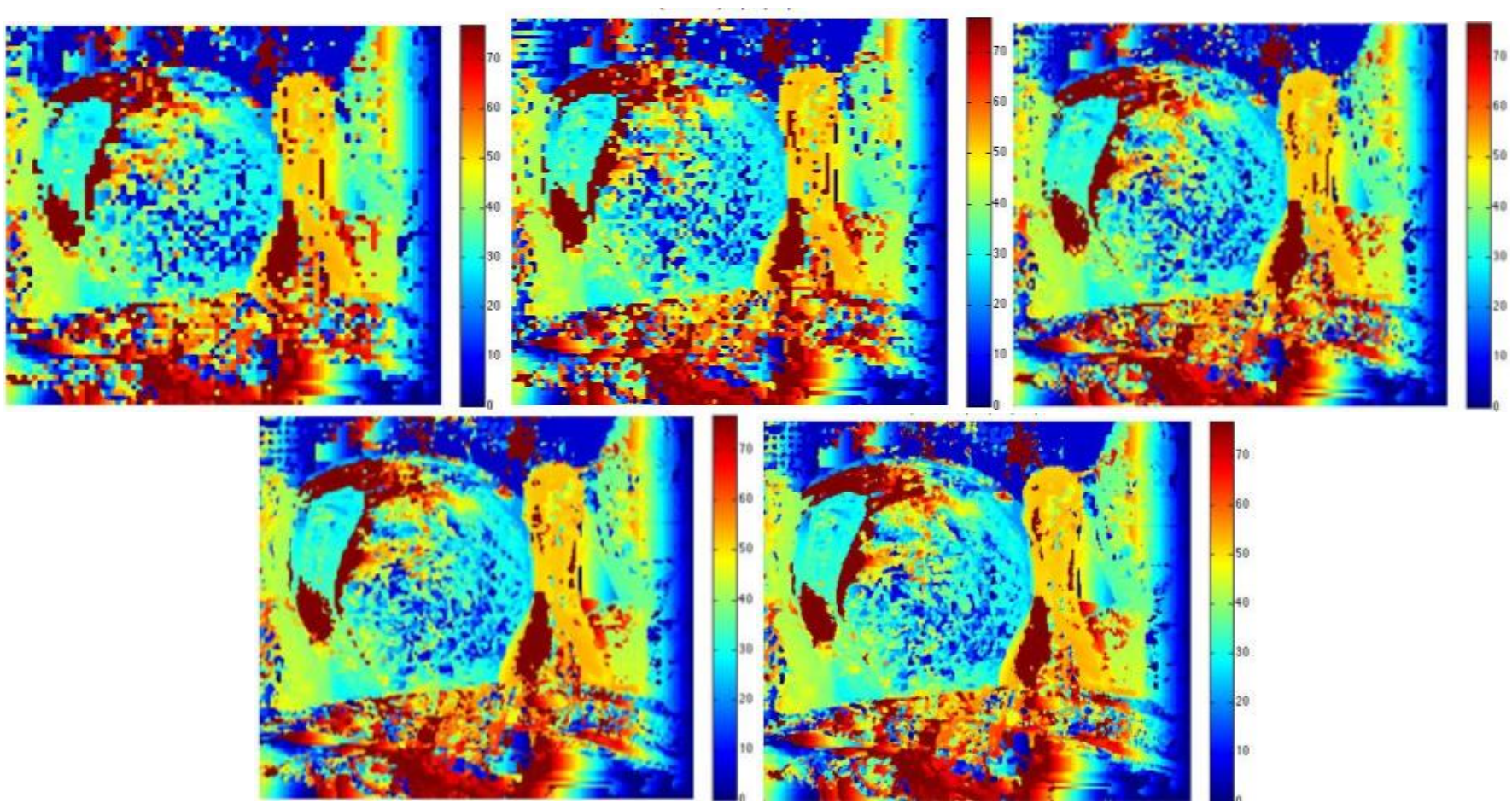

Figure 16: X disparity map for with different overlaps for a window size of 5x5 pixels (Dataset: Bowlign1)

The figure below shows the disparity map in the y direction for the 'Bowling1' dataset for a window size of 5x5 pixels and 3 different overlap settings of 2, 3 and 4 pixels. The images in the Middlebury dataset are rectified but figure 17 shows that there are some variations in the y direction which were not accounted for in the rectification process. This may be due to pixel occlusions [17]. The advantage of performing a coarse alignment in the analog domain is that the operation is being performed on raw image data i.e. unrectified images. Hence we calculate disparities in both x and y directions. This gives us extra information about the pixel disparities which cannot be obtained by performing stereo correspondence on rectified stereo images.

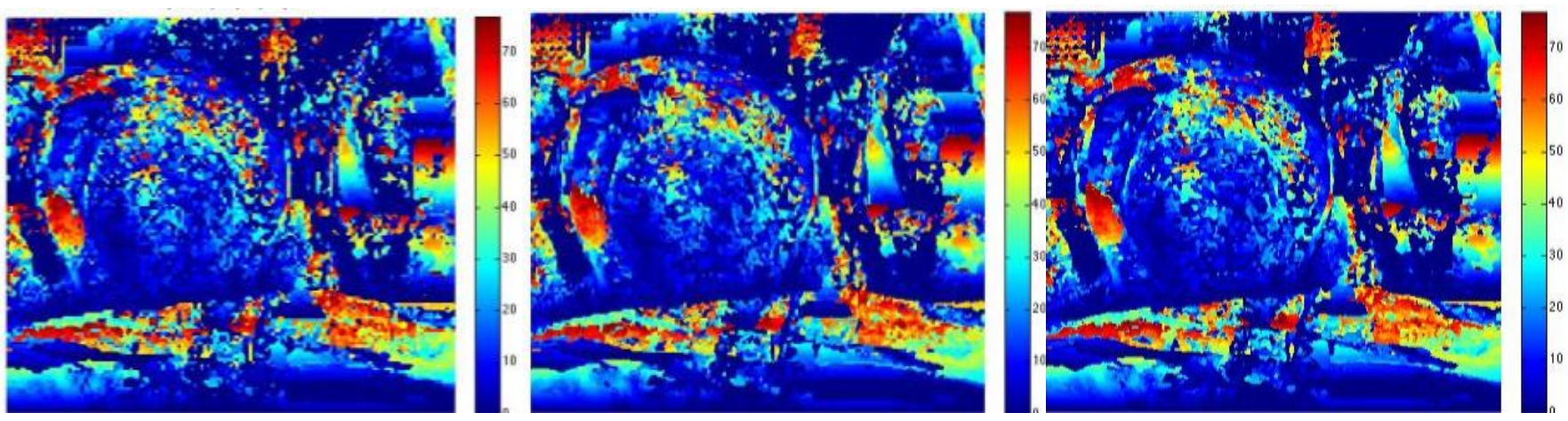

Figure 17: Y disparity map for with different overlaps for a window size of 5x5 pixels (Dataset: 'Bowling1')

## 4. CONCLUSIONS

In this work we propose analog signal processing as a solution for handling the high computational load of some of the image processing algorithms while simultaneously meeting the reduced SWaP requirements. The analog processor will be used to augment the digital processor and work in parallel with it to perform key computations, making the system faster and more efficient. We implement two highly computational stereo correspondence algorithms to align stereo image pairs. Novel modifications were proposed to both the NCC algorithm and SAD algorithm which reduced the computation and made the algorithm efficiently implementable in analog hardware. The modified NCC algorithm has a 50% reduction in Matlab computation time over the original algorithm and the modified SAD algorithm has a speed up of more than 4x in Matlab computation time over the original algorithm. An approximate power consumption of 215.15mW was obtained for the analog correlation block. The SAD algorithm produced a finer depth map. The actual analog hardware implementation of the algorithm and the new imaging architecture will contribute to a further reduction in computation time as compared to a digital implementation. Various other simulations were also run to check the robustness and performance of the algorithm. The experimental results obtained are very promising and we believe analog processing will be a viable solution to these problems. As a part of the future work and as a proof-of-concept the analog image correlator circuit will be built from commercially available off the shelf components. A test plan will be setup for this circuit. Once the required results are obtained, the next step will be to build the architecture in silicon.

## Authors

**Nihar Athreyas** received his B.E. degree in Electronics and Communication Engineering from VTU Belgaum, India and M.S. degree in Electrical and Computer Engineering from University of Massachusetts, Amherst, MA, in 2010 and 2013 respectively. He is currently pursuing his doctoral degree under the supervision of Dr. Dev Gupta at University of Massachusetts, Amherst, MA. He joined Newlans, Inc., Acton, MA in June of 2014 as an Intern. His current research interests include communications, CMOS analog design and applications of analog signal processing.

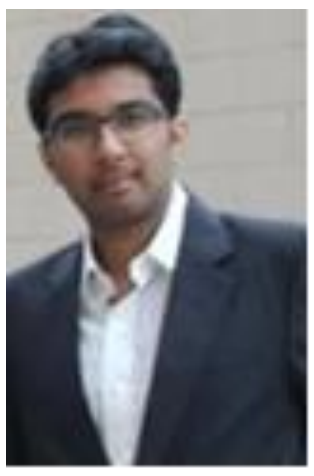

**Zhiguo Lai** received the B.S. degree in mechanical engineering from Tsinghua University, Beijing, China, the M.S. degree in electrical and computer engineering from University of Alaska, Fairbanks, AK, and the Ph.D. degree in electrical engineering from University of Massachusetts, Amherst, MA, in 1999, 2002, and 2007, respectively. From summer of 2007 to summer of 2009, he was a postdoctoral fellow at University of Massachusetts, Amherst, MA, working on narrowband interference mitigation for UWB systems. Since 2009, he has been with Newlans, Inc., Acton, MA. His research interests include design of programmable CMOS filters, wideband analog signal processing, and quantum computation. He is currently working on analog computation using deep-submicron CMOS technology.

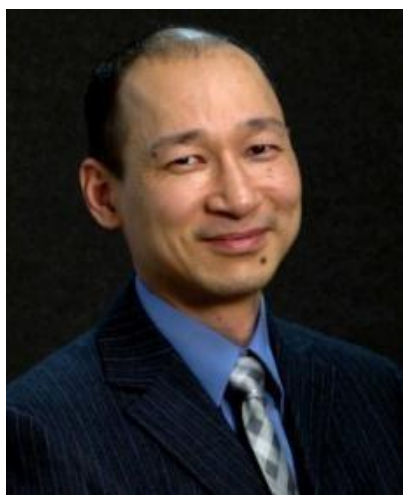

**Dev V. Gupta** received his Ph.D. in 1977 from the University of Massachusetts, Amherst. He held various engineering positions at the Bell Laboratories in Andover, MA, from 1977 to 1985. He was the General Manager at Integrated Network Corporation, a manufacturer of DSL access products, from 1985 to 1995.

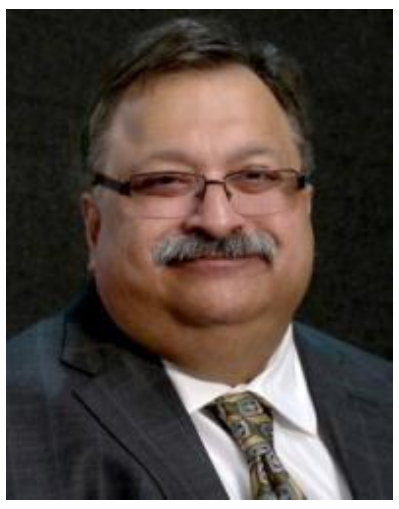

Dr. Gupta founded two companies, Dagaz Technologies and Maxcomm, which were acquired by Cisco Systems in 1997 and 1999 respectively. These companies developed and manufactured telephone exchange and voice and data equipment for DSL. He was Cisco's VP of Architecture in the access business unit between founding Dagaz Technologies and Maxcomm. In 2000, he founded Narad Networks which manufactured Gigabit Ethernet networking equipment for the cable industry. Narad Networks (renamed PhyFlex) was acquired by Cienna in 2007. Newlans was founded in 2003. He is a Charter Member of the Atlantic chapter of the Indus Entrepreneurs (TIE), an organization which promotes entrepreneurship. The World Economic Forum named him a 'Tech Pioneer' for the years 2001 and 2002. He is a Trustee of the University of Massachusetts, Amherst and a board member of the UMass Foundation. He is an Adjunct Professor at the University of Massachusetts where he created the Gupta Chair in the Department of Electrical and Computer Engineering. He has over thirty patents in communications, networking, circuit design, and signal processing. Dr. Gupta is the President and CEO of Newlans and he provides technical vision for the Company.

**Jai Gupta** currently serves as Chief Architect for commercialization efforts of Newlans' analog integrated circuits and wideband signal processing technology. His primary role is system architecture design and program management. He has previously held research positions at the National Institute of Standards and Technology and Huntington Medical Research Institute, worked as a systems engineer at metropolitan ad-hoc networking startup Windspeed Access, and held internship positions at broadband cable startup Narad Networks as well as Cisco Systems.
Jai recently completed the Executive MBA degree at Duke University's Fuqua School of Business with a concentration in Entrepreneurship and Innovation. He previously obtained the MSEE degree in Electrical Engineering Systems from the University of Southern California's Viterbi School of Engineering and the BSEE degree from the University of Pennsylvania's Moore School of Electrical Engineering.